\documentclass[conference]{IEEEtran}
\IEEEoverridecommandlockouts
\usepackage{cite}
\usepackage{amsmath,amssymb,amsfonts}
\usepackage{graphicx}
\usepackage{textcomp}
\usepackage{marginnote} 
\usepackage[flushleft]{threeparttable}
\usepackage{xcolor}
\def\BibTeX{{\rm B\kern-.05em{\sc i\kern-.025em b}\kern-.08em
    T\kern-.1667em\lower.7ex\hbox{E}\kern-.125emX}}
\usepackage{tabularx}
\usepackage{subcaption}
\usepackage{url}
\usepackage{hyperref}
\usepackage{soul}
\usepackage{balance}
\usepackage{bm}% added by zsh
\usepackage{algorithm} 
\usepackage{eucal}
\usepackage[algo2e]{algorithm2e} 
\usepackage{algpseudocode}
\usepackage{booktabs} 
\usepackage[usestackEOL]{stackengine}
\usepackage{adjustbox}
\usepackage{array}
\usepackage{multirow}

\begin{document}

\title{ResLearn: Transformer-based Residual Learning for Metaverse Network Traffic Prediction
%\thanks{Funding placeholder}
%\thanks{(\textit{Corresponding author: }, e-mail: )}
}

\author{\IEEEauthorblockN{Yoga Suhas Kuruba Manjunath\IEEEauthorrefmark{1}, Mathew Szymanowski \IEEEauthorrefmark{1},
Austin Wissborn\IEEEauthorrefmark{1},\\Mushu Li \IEEEauthorrefmark{2}, Lian Zhao\IEEEauthorrefmark{1},  and Xiao-Ping Zhang \IEEEauthorrefmark{3}}
\IEEEauthorblockA{\IEEEauthorrefmark{1}Department of Electrical, Computer \& Biomedical Engineering, Toronto Metropolitan University, Toronto, Canada\\
\IEEEauthorrefmark{2}Department of Computer Science and Engineering, Lehigh University, Bethlehem, PA, USA\\
\IEEEauthorrefmark{3} Shenzhen Key Laboratory of Ubiquitous Data Enabling, Tsinghua Shenzhen International \\Graduate School, Tsinghua University, \\%and with the Department of Electrical, Computer and Biomedical Engineering, \\Toronto Metropolitan University, Toronto, ON M5B 2K3, Canada.\\
Email: \{yoga.kuruba@torontomu.ca, austin.wissborn@torontomu.ca, mszymanowski@torontomu.ca,\\ mul224@lehigh.edu, l5zhao@torontomu.ca, xpzhang@ieee.org\}}}

\maketitle

\begin{abstract}
Our work proposes a comprehensive solution for predicting Metaverse network traffic, addressing the growing demand for intelligent resource management in eXtended Reality (XR) services. We first introduce a state-of-the-art testbed capturing a real-world dataset of virtual reality (VR), augmented reality (AR), and mixed reality (MR) traffic, made openly available for further research. To enhance prediction accuracy, we then propose a novel view-frame (VF) algorithm that accurately identifies video frames from traffic while ensuring privacy compliance, and we develop a Transformer-based progressive error-learning algorithm, referred to as \textit{ResLearn} for Metaverse traffic prediction. ResLearn significantly improves time-series predictions by using fully connected neural networks to reduce errors, particularly during peak traffic, outperforming prior work by 99\%. Our contributions offer Internet service providers (ISPs) robust tools for real-time network management to satisfy Quality of Service (QoS) and enhance user experience in the Metaverse.

\end{abstract}


\section{Introduction}

% Story for the introduction

% First paragrah
% Brief introduction on Metaverse and different realitites. Use some good diagram
% Increase in the popularity of Metaverse Services

% second paragraph
% problem with increase in Metaverse traffic
% effects of inefficiency of resource management on QoE
% Why it is required to predict Metaverse network traffic

% third and/or 4th paragraph - Realted works
% Use atleast 6 tp 7 latest works in metaverse network traffic prediction
% identify problems in them to find research gap
% For you Reference: Metaverse traffic includes: VR, AR, MR. Also, you can use some video network traffic prediction as well.

% v3
The Metaverse is a comprehensive ecosystem of interconnected virtual worlds that provide immersive experiences to users. The ecosystem enhances existing and generates new value from economic, environmental, social, and cultural perspectives~\cite{kontogianni2024towards}. Services in the Metaverse ecosystem are designed to be accessed using immersive extended reality (XR) environments. XR is an umbrella term that describes the technologies affecting the user's immersive experience, such as virtual reality (VR), augmented reality (AR), and mixed reality (MR)~\cite{suh2018state}. VR allows users to interact with virtually generated environments designed to simulate real-world experiences. AR overlays interactive, virtually generated information onto real-world objects or within real-world spaces. XR technologies lie on a spectrum between AR and VR. In cases where the distinction between the realities is ambiguous, the experience is considered MR. As the Metaverse's growth continues and XR evolves, the popularity of its services increases. Driven by the rapid growth of the Metaverse, Internet traffic is expected to surpass current forecasts significantly ~\cite{ericsson2022}. The entertainment and social media industries have seen the most substantial growth of Metaverse services, as evidenced by popular virtual performance events, one of which attracted an audience of 36 million users \cite{qualcomm2020, roblox2021}. Healthcare, training, and marketing for Metaverse services have also grown recently~\cite{musamih2022metaverse, wang2023survey}. Cloud rendering for Metaverse is crucial to offloading computing resources to make the services affordable, a popular technique for VR games ~\cite{zhao2021virtual}. Consequently, the Ericsson 2022 report emphasizes the growing need for more intelligent interactions between XR services and the network to maintain high Quality of Service (QoS)~\cite{ericsson2022}. Therefore, network management is crucial for Internet service providers (ISPs) to accommodate adequate resources and avoid cybersickness among users~\cite{feldstein2020simple, 10124955, 10673991}.

Metaverse traffic consists of video, audio, and control flows \cite{zhao2021virtual}, among all downlink video frames are resource-demanding in the case of VR, and both uplink/downlink video frames for AR and MR traffic. Therefore, predicting the frame size is vital, and for latency-related issues, it is essential to predict frame inter-arrival time and frame duration \cite{9622181}. Predicting frame size, inter-arrival time, and frame duration will help ISPs prepare and manage the Metaverse network for holistic and intelligent traffic management. However, there needs to be more real-world Metaverse data and research in prediction to make progress in the field. Essentially, frame-related information is time series data. The recent advent of different state-of-the-art artificial intelligence (AI) based time series models has shown tremendous progress in time series predictions \cite{jin2024survey}. The only VR frame size prediction work is available at \cite{10437897}; therefore, we consider it state-of-the-art (SoA) work for benchmark comparison. The work studies different AI models to predict VR frame size. It establishes stacked LSTM to produce better results based on transfer learning methodologies. Therefore, the solution aims for online prediction, imperative for real-time network management. However, the work is evaluated on a small dataset captured in a controlled environment. Also, the performance can be improved with further reduction in the error. The frame identification methodology used in the work might need to be revised because the frame loss for 120 Mbps is more than 54 Mbps since the dataset is captured in a controlled environment. Our literature review identifies the following gaps: i) the need for holistic, comprehensive, and real-world Metaverse datasets and ii) accurate frame-related data predicting algorithms.

We chart out a state-of-the-art Metaverse testbed to capture a holistic, comprehensive, and real-world Metaverse dataset comprising VR games, VR videos, VR chat/VoIP, AR, and MR traffic. We also make it open for the research fraternity available at \cite{data_1}. Our work treats the Metaverse traffic in segments to enhance the predictability of the data at higher speed. We propose an accurate view-frame (VF) algorithm that helps to identify the types of video frames using application-level information to comply with privacy-related policies. Our proposed VF algorithm can predict the number of frames in a segment, total frame size, and frame inter-arrival time. Finally, we propose a state-of-the-art residual learning (ResLearn) algorithm that uses transformer \cite{vaswani2017attention} to predict time-series data and fully connected neural networks (FCNN) to learn errors with a bias to identify the peaks of the frame-related data for accurate network management. Our solution outperforms the SoA work \cite{10437897} by 99\% in reducing the prediction errors. The implementation of the solution is made open for the research community at \cite{yogasuha18}. %Sections detailing our solution are aligned with the rest of the paper.

\section{System Model}

\begin{figure}
	\centering
	\includegraphics[width=0.8\linewidth]{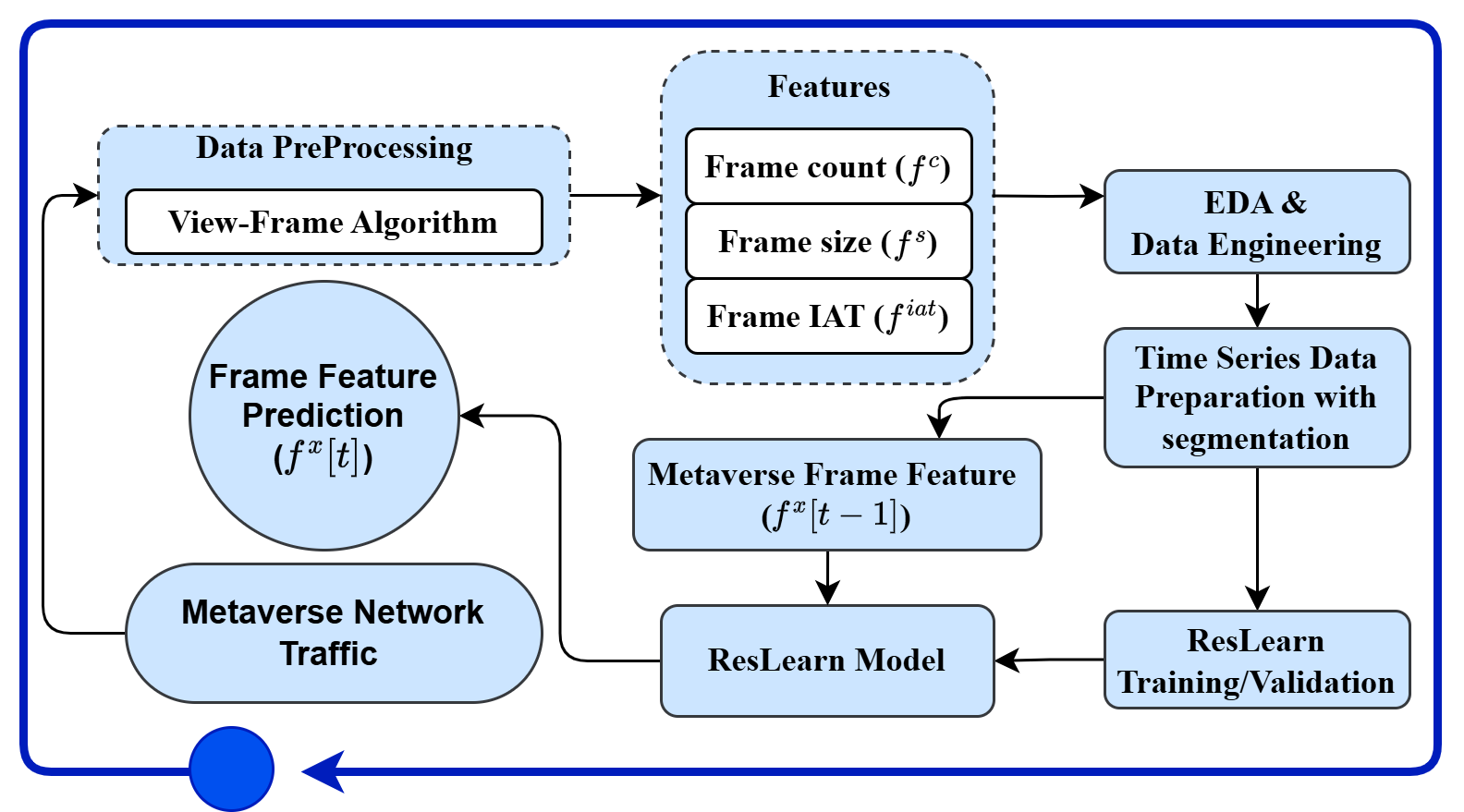}
	%\vspace{-0.2cm}
	\caption{System model of the proposed solution. $x$ in $f^x[t]$ represents one of the features. EDA is the exploratory data analysis.}
	\label{fig:sm}
	\vspace{-0.5cm}
\end{figure}

% \begin{figure}
% %\vspace{-0.4cm}
% %\title{}
% \centering
% \subfloat[]{%
%   \includegraphics[width=0.8\linewidth]{reslearn.png}%
%   \label{fig:sm_1}
% }
% \vspace{-0.01cm}
% \subfloat[]{
% \includegraphics[trim=0 0 0.3cm 0 ,clip, width=0.8\linewidth]{reslearn_1.png}%
% \label{fig:sm_2}
% }
% %\vspace{0.1cm}
% \vspace{-0.1cm}
% \caption{(a) System model of the proposed solution, and (b) Residual Learning (ResLearn) algorithm's training process.}
% \vspace{-1cm}
% \label{fig:sm}
% \end{figure}

\begin{figure}
%\vspace{-0.4cm}
%\title{}
\centering
\subfloat[]{%
  \includegraphics[width=0.8\linewidth]{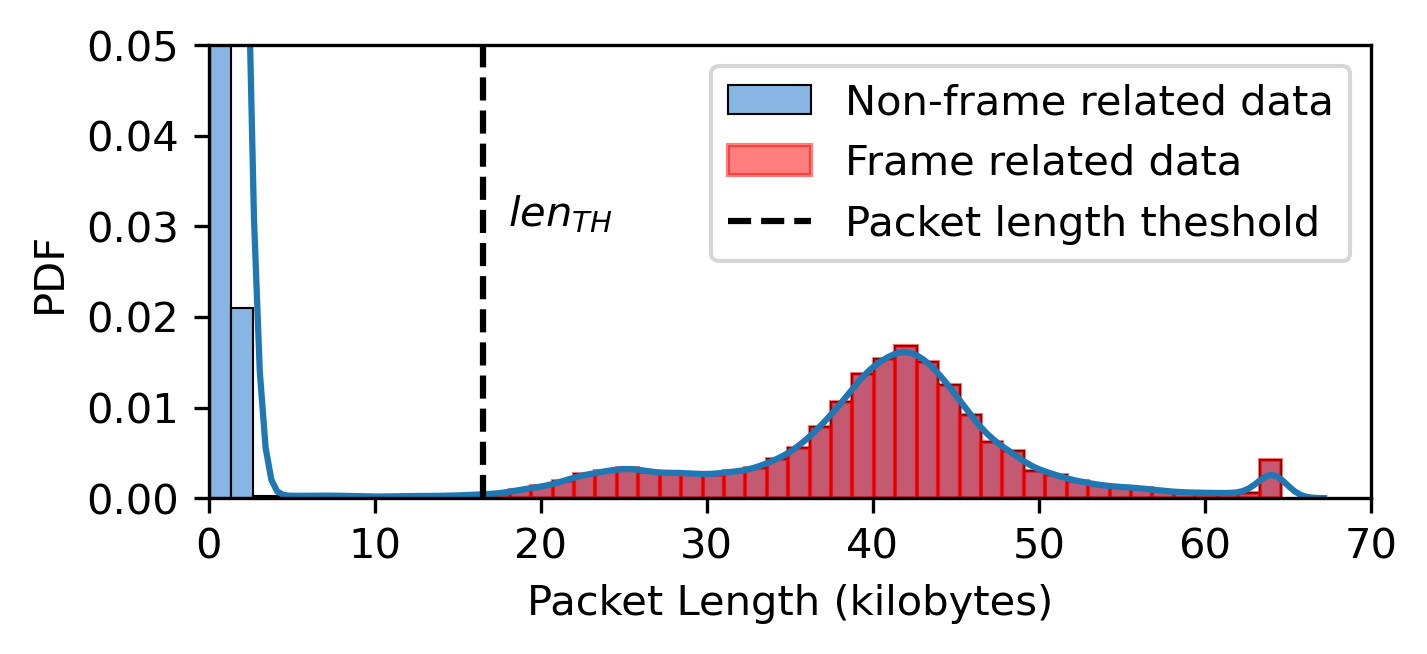}%
  \label{fig:iat+pktlen(a)}
}
\vspace{-0.01cm}
\subfloat[]{
\includegraphics[width=0.8\linewidth]{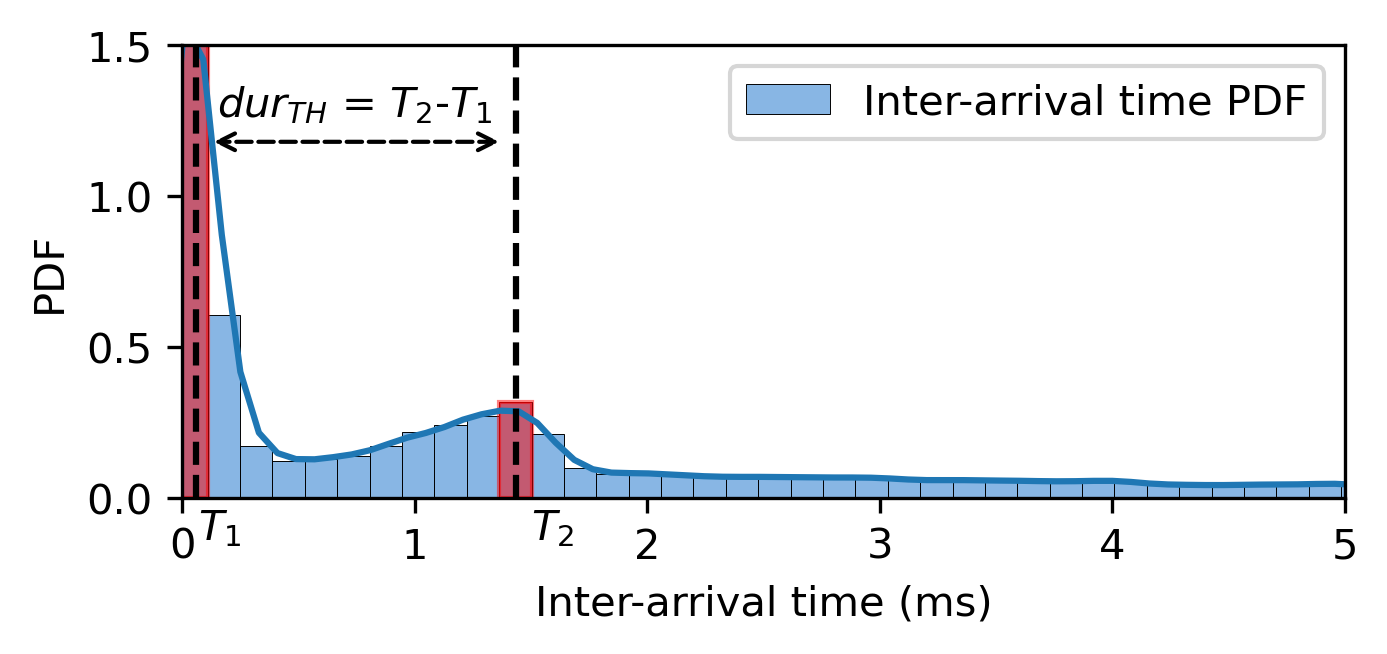}%
\label{fig:iat+pktlen(b)}
}
%\vspace{0.1cm}
%\vspace{-0.1cm}
\caption{(a) PDF of packet lengths, and (b) PDF of inter-arrival time for a sample Metaverse traffic segment.%\\ 
%$len_{TH}$ is determined as 25\% of the maximum length of the observed packet length. $dur_{th}$ is the frame duration threshold determined between the first two peaks. The first peak represents the start of the video frame packet with less inter-arrival time, and the second peak represents the end of the video frame when audio and control-related flows are transmitted with slightly higher inter-arrival time. The $len_{TH}$ and $dur_{th}$ are calculated from the first segment at a session's start for the VF algorithm. All packets within these thresholds are collected to identify the frames. The method is verified on different Metaverse rendering platforms: Meta air link \cite{airlink}, and Virtual Desktop Streamer (VDS) \cite{HomeVirt30}.}
}
\vspace{-0.65cm}
\label{fig:iat+pktlen}
\end{figure}

The system model, in Figure \ref{fig:sm} illustrates the proposed framework for predicting Metaverse network traffic, focusing on frame-level features. The process begins with data preprocessing, where the VF algorithm is applied to extract relevant frame-related data. The key features extracted from the incoming traffic include frame count ($f^c$), frame size ($f^s$), and frame inter-arrival time (IAT) ($f^{iat}$). Our VF algorithm works on application-level features: time, packet length, packet direction, and packet inter-arrival time. Frame-related packets have a more considerable length with relatively more minor inter-arrival time. We use this property to determine the thresholds for packet length and inter-arrival time to identify frame-related packets as shown in Figure \ref{fig:iat+pktlen}. In Figure \ref{fig:iat+pktlen}, $len_{TH}$ is determined as 25\% of the maximum length of the observed packet length. $dur_{th}$ is the frame duration threshold determined between the first two peaks. The first peak represents the start of the video frame packet with less inter-arrival time, and the second peak represents the end of the video frame when audio and control-related flows are transmitted with slightly higher inter-arrival time. The $len_{TH}$ and $dur_{th}$ are calculated from the first segment at a session's start for the VF algorithm. All packets within these thresholds are collected to identify the frames. The method is verified on different Metaverse rendering platforms: Meta air link \cite{airlink}, and Virtual Desktop Streamer (VDS) \cite{HomeVirt30}.

Based on the requirement, one of the features is selected for prediction. These features undergo Exploratory Data Analysis (EDA) and Data Engineering to ensure data quality and integrity, followed by Time Series Data Preparation with segmentation for model training. The core of the system is frame feature prediction ($f^x[t]$), where the previous frame feature values ($f^x[t-1]$) are used as inputs to predict the current network traffic behaviour. With this information, we provide the mathematical problem statement to predict key features of the Metaverse network traffic features. Let $\bm{f}_t$ be a frame vector, identified by VF algorithm for a given segment, having frame-related information given as $\bm{f}_t =[f^c_t, f^s_t, f^{iat}_t],$ where $t$ indicates the time index of the given segment. Let $f^x[t-1]$ represents the Metaverse frame-related network traffic at the previous time step, where $x$ indicates one of the three frame-related features in $\bm{f}_{[t-1]}$. Therefore, frame count, size, and inter-arrival time are individually predicted using historical data. The prediction function for the current network traffic, $f^x[t]$, is 

\begin{equation}
     f^x[t] = \psi(f^x[t-1]),
     \label{eq:1}
\end{equation} 

where $\psi(\cdot)$ is the prediction model. %that can be obtained using proposed ResLearn algorithm, detailed in Section III.

%We perform online training in segments from which we use part of the data from a segment for training and the rest for validation. The segment size is $N$; we use 0.6*($N$) for training and 0.4*($N$) for validation. 

%The core of the system is frame feature prediction ($f^x[t]$), where the previous frame feature values ($f^x[t-1]$) are used as inputs to predict the current network traffic behaviour. 

%This prediction task is managed using a ResLearn algorithm, a residual learning-based approach to handle time-dependent data. The model in ResLearn algorithm, i.e., ResLearn model, is trained and validated iteratively, adjusting to traffic fluctuations over time. The predicted frame features feed back into the system, contributing to the overall Metaverse network traffic estimation. Finally, the model is adopted for real-time prediction as the validation error decreases and stabilizes. With this process, ResLearn mainly learns the long-term and short-term dependencies and progressive errors; therefore, the deployed model can properly increase the prediction accuracy if there is no behavioural change, where Metaverse traffic usually follows common distributions showing small to no behavioural change \cite{zhao2021virtual}. The ResLearn algorithm's training process is shown in Figure \ref{fig:sm_2}. This cyclical process allows for real-time and accurate traffic predictions, which is crucial for maintaining seamless network performance in Metaverse environments.
%\section{Residual Learning}

\section{ResLearn: Transformer-based Residual Learning for Metaverse Network Traffic Prediction}

The residual Learning (ResLearn) algorithm is a two-step prediction approach involving a transformer deep neural network model \cite{vaswani2017attention} designed to enhance Metaverse network traffic forecasting by leveraging residual learning inspired by ResNet \cite{7780459}. Transformer is known for learning short and long-term dependencies from time series data. However, the prediction error is inevitable due to the randomness introduced by network health and users in Metaverse infrastructure. However, we can learn the nature of error using a neural network. ResLearn is a novel approach that uses a Transformer in the first step, given as $\mathcal{F}_1(\cdot)$, the predictive Transformer model. The residual from the  $\mathcal{F}_1(\cdot)$ is fed to a fully connected neural network (FCNN) to learn the nature of the error, given as $\mathcal{F}_2(\cdot)$. The final prediction model from Eq. \eqref{eq:1} is given as $\psi(\cdot) = \mathcal{F}_1(\cdot) + \mathcal{F}_2(\cdot)$.

\begin{figure}
	\centering
	\includegraphics[width=0.8\linewidth]{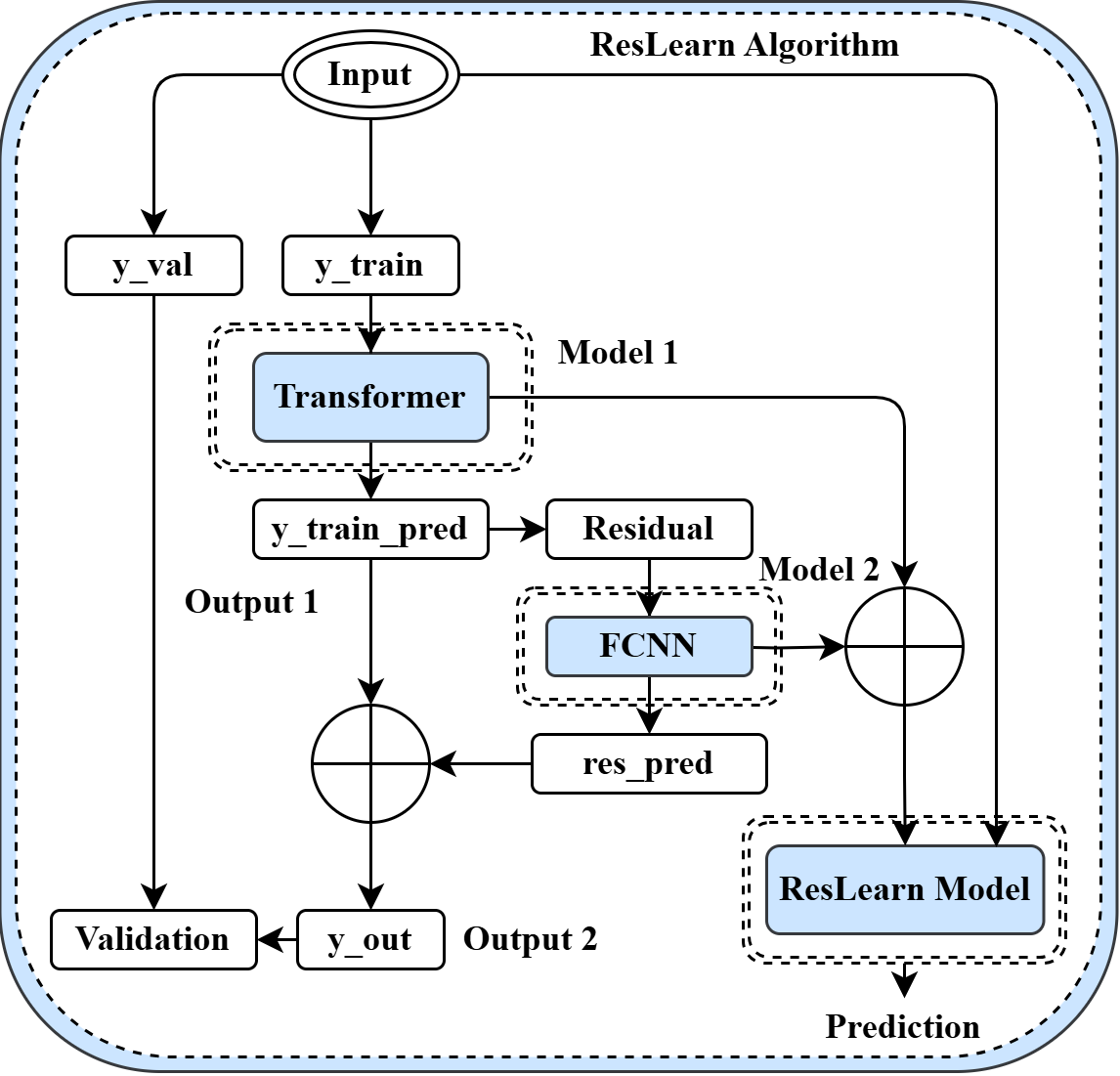}
	%\vspace{-0.2cm}
	\caption{ResLearn algorithm.}
	\label{fig:rl_3}
	%\vspace{-0.5cm}
\end{figure}

Figure \ref{fig:rl_3} illustrates the workflow of the ResLearn algorithm.
The input data is split into training ($y_\text{train}$) and validation ($y_\text{val}$) sets. The first model, including a transformer network, processes the training data and generates an initial prediction result ($y_\text{train\_pred}$).  This output, called Output 1, is then compared to the actual training data to calculate the residuals, which capture the difference between the predicted and true values. These residuals are passed to the second model, an FCNN designed to learn and predict the patterns in the residuals, producing a corrected prediction ($res_\text{pred}$). Both predictions from the transformer and FCNN are combined to form the final output ($y_\text{out}$) of the ResLearn model. This combined output is then validated with the unseen validation data, improving the final prediction's overall accuracy. The ResLearn training algorithm is shown in Algorithm \ref{Alg:rs}. $S_N^X$ is a time-series data segment, where $N$ is the segment size and $X$ is the number of segments. For each segment, the data is split into training and validation sets. The transformer model is trained on the training set, and its predictions, $T_{PR}$, are computed. The residuals, representing the error between the actual and predicted values, are calculated and adjusted by adding a bias, $Res_B$, to highlight essential peaks. The dense model is then trained on these residuals, and the final ResLearn model, $M_{RL}$, is created by combining the outputs of both models. The combined model is then evaluated on the validation set using error metrics. The process is repeated for each segment, refining the prediction accuracy iteratively until the validation error is stabilized for $M_{RL}$. The algorithm's output is the final ResLearn model $M_{RL}$, which integrates the strengths of both models to deliver improved predictive performance. 

\begin{algorithm}
\caption{Residual Learning (ResLearn) Training}\label{alg:Reslearn}
\KwData{Time series segmented data ($S_N^X \in S_N$) where $N$ is the size of the segment and $X$ is segment number}
\KwResult{ResLearn model ($M_{RL}$).}
\textbf{initialization}\;
 
 %$\bm{S}$ = $[$ $]\;$ %\Comment{Empty array to form matrix}
%$Z_{error}$ = $0$\;;$E_{stop}$ = $0$\;;models = $[$ $]\;$;data = $[$ $]\;$

model1 = transformer() \Comment{to learn short and long-term dependencies}

model2 = dense() \Comment{to learn residuals}
  
\While{$i$ != $X$}{
    %$X_{train}$, $y_{train}$,$X_{val}$, $y_{val}$ = %splitData($S_X_{i}^N$, $T_{R}$)\;

    train, val = splitData($S_{N_{i}}^X$, $T_{R}$)\;
    
    model1.Train(train)

    $T_{PR}$ = model1.Predict(train) \Comment{predicts the data}

    $Res$ = train - $T_{PR}$

    $Res_{B}$ = $|min(Res)|$ \Comment{bias to identify peaks}

    residualTrain = $Res$ + $Res_{B}$

    model2.Train(residualTrain)

    $M_{RL}$ = model1 + model2

    $Res_{PR}$ = $M_{RL}$.Predict(val) 

    error\_metrics(val, $Res_{PR}$) \Comment{calculates all error metrics}

    i++
}
  %\vspace{-0.1cm}
\Return {$M_{RL}$}

\label{Alg:rs}
\end{algorithm}

%It begins by segmenting the time-series data into smaller portions, denoted as $S_N^X$, where $N$ is the segment size and $X$ is the number of segments. The algorithm initializes two models: the first, a transformer model, captures both short-term and long-term dependencies in the data, while the second, a Dense model, is responsible for learning from the residuals. For each segment, the data is split into training and validation sets. The transformer model is trained on the training set, and its predictions, $T_{PR}$, are computed. The residuals, representing the error between the actual and predicted values, are calculated and adjusted by adding a bias, $Res_B$, to highlight essential peaks. The dense model is then trained on these residuals, and the final ResLearn model, $M_{RL}$, is created by combining the outputs of both models. The combined model is then evaluated on the validation set using error metrics, and the process is repeated for each segment, refining the prediction accuracy iteratively. The algorithm's output is the final ResLearn model $M_{RL}$, which integrates the strengths of both models to deliver improved predictive performance.

The time complexity of the ResLearn algorithm, which involves training a transformer model and a FCNN, can be approximated as follows: the training complexity of the transformer model is typically $O(T \cdot N^2)$, where $T$ is the number of training epochs and $N$ is the sequence length. The training complexity of the FCNN can be approximated as $O(T' \cdot N \cdot D)$, where $T'$ is the number of epochs, $D$ is the number of neurons in the hidden layer, and $N$ represents the number of training samples \cite{shah2022time}. Considering $X$ segments of data, the overall time complexity of the algorithm is given by $O(X \cdot (T \cdot N^2 + T' \cdot N \cdot D)).$

%In summary, the total time complexity for the ResLearn algorithm is approximately 

%$O(X \cdot T \cdot N^2 + X \cdot T' \cdot N \cdot D).$

\section{Experimentation Setup and Results}
\subsection{Datasets and predictability}
%\subsection{Data predictability}

%\subsection{Implementation and Plan of Experiments}

%\usepackage{multirow}

\begin{figure}
	\centering
	\includegraphics[width=0.9\linewidth]{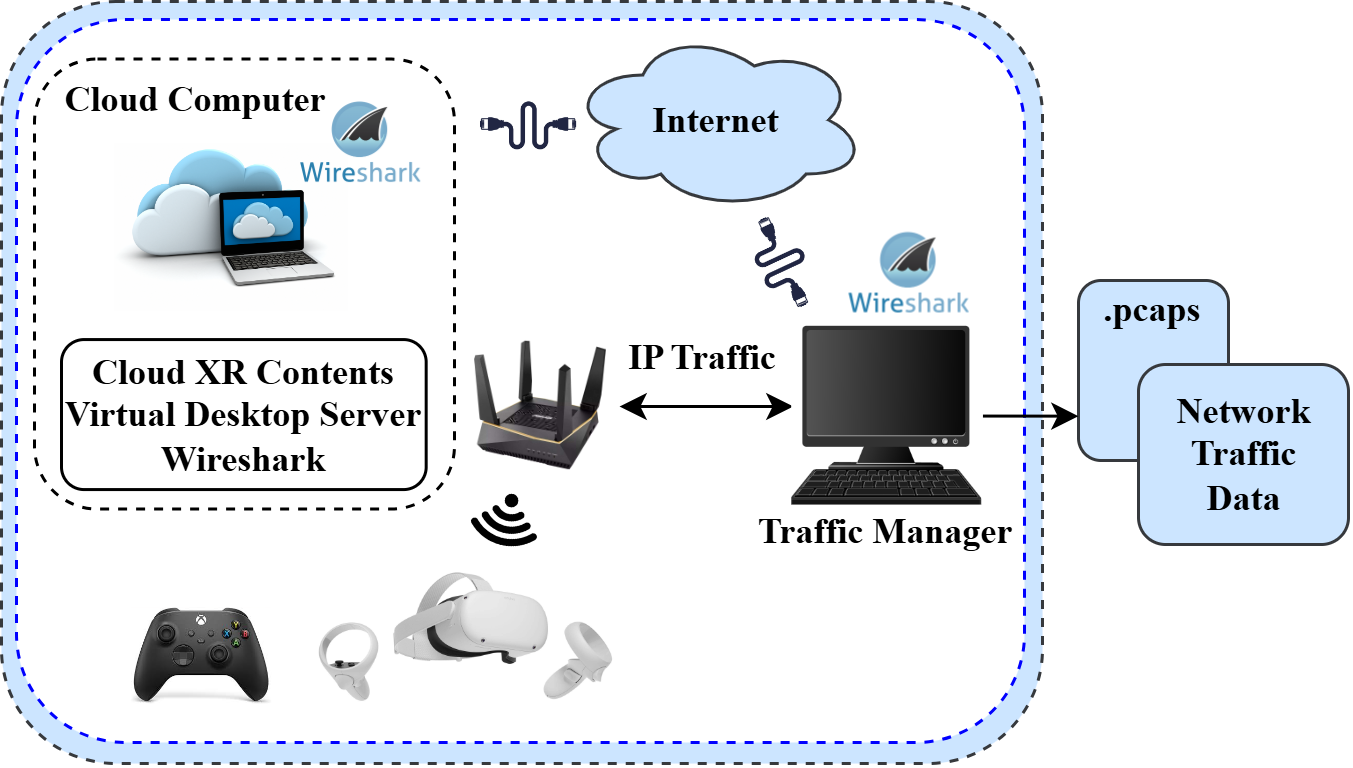}
	%\vspace{-0.2cm}
	\caption{Experimental platform used for data capture.}
	\label{fig:tb}
	%\vspace{-0.5cm}
\end{figure}

The experiments are designed to evaluate the performance of various VR, AR, and MR services across three datasets, as shown in Table \ref{table:exp}. The Dataset I \cite{data_1} (in-house dataset) focuses on diverse services, including gaming, video streaming, and communication (chat/VoIP), and is tested with applications such as Dirt Rally 2.0, Bigscreen, VR Chat, Solar System, and Reality Mixer. Figure \ref{fig:tb} shows the testbed used in the data capture. In the testbed, a virtual desktop streamer (VDS) rendering platform is used for the setup. A cloud computer with a VDS server is a rendering device to which the VDS client on the Oculus Quest 2 is connected. Traffic manager is used to simulate low latency networks to replicate real-world scenarios. Traffic is captured on the cloud computer using Wireshark. More details and packet captures (pcap) are available at \cite{data_1}. Dataset II \cite{zhao2021virtual} examines slow and fast VR traffic using Steam VR Home and Beat Saber to study the impact of different traffic patterns. Dataset III \cite{questset} involves two experiments: the first explores fast and slow VR traffic across Beat Saber, Medal of Honor, Forklift Simulator, and Cooking Simulator, while the second focuses on a subset of applications (Forklift Simulator, Cooking Simulator, Beat Saber, and Medal of Honor) to assess network performance under varying traffic conditions further. We use 50\% of data for training, in which 20\% of the data is used for validation. Another 50\% of the data is used for testing. The solution is developed in Python using data science libraries such as Scikit-learn (Sklearn), NumPy, and Pandas. The implementation is available at \cite{yogasuha18}. The experiments are conducted on a Windows system with an Nvidia RTX2800S GPU. The Windows environment is set up with Anaconda to support machine learning libraries, including TensorFlow.

\begin{figure}
	\centering
	\includegraphics[trim={2 1.5 2 1.5},clip, width=\linewidth]{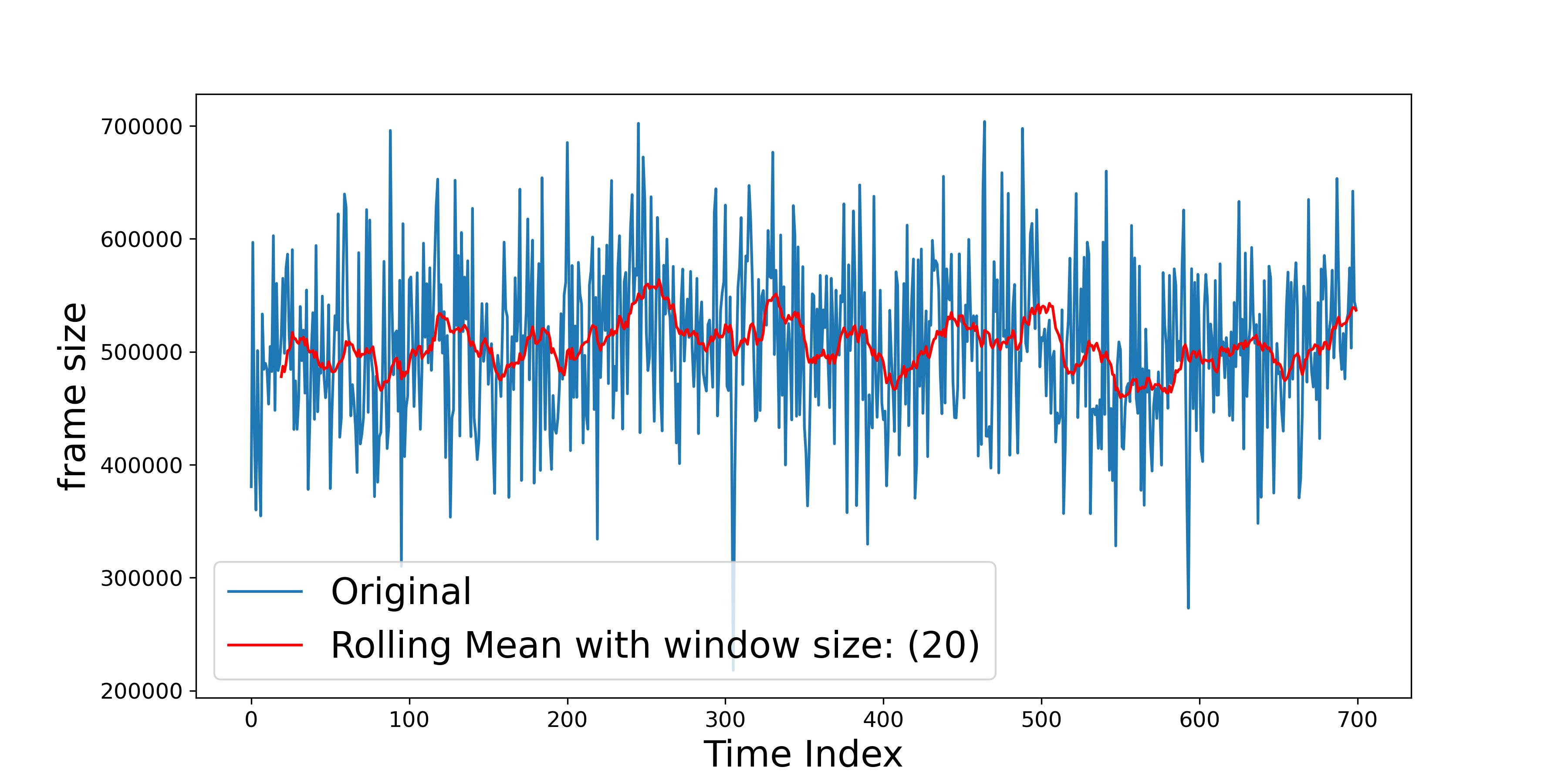}p
	%\vspace{-0.2cm}
	\caption{frame size time series data with roll-over averaging window.}
	\label{fig:rs}
	%\vspace{-0.5cm}
\end{figure}

The analysis of data predictability begins with an exploratory data analysis (EDA) that suggests randomness, shown in Figure \ref{fig:rs} depicted by blue line, in the network traffic data. Runs test is a statistical procedure which determines whether a sequence of data within a given distribution have been derived with a random process or not \cite{asano1965runs}. Runs test of the raw data provides a p-value of 0.15, which indicates no significant evidence against the null hypothesis of randomness. However, a deeper examination using decomposition techniques reveals underlying patterns in the time series, breaking it down into trend, seasonal, and residual components. This is further reinforced by applying a rolling window average (with a window size of 20), where the rolling mean (depicted by the red line in Figure \ref{fig:rs}) closely follows the shape of the data, revealing a clear trend. The corresponding Runs test, now with a p-value close to zero, confirms that the data is predictable and not random. Rolling statistics effectively uncovers this structure, making the data suitable for forecasting models. %also corroborated by the autocorrelation graphs shown in figure \ref{fig:cor}.

% \begin{figure}
% %\vspace{-0.4cm}
% %\title{}
% \centering
% \subfloat[]{%
%   \includegraphics[width=0.8\linewidth]{raw_acor.png}%
%   %\label{fig:iat+pktlen(a)}
% }
% \vspace{-0.01cm}
% \subfloat[]{
% \includegraphics[width=0.8\linewidth]{avg_acor.png}%
% %\label{fig:iat+pktlen(b)}
% }
% %\vspace{0.1cm}
% %\vspace{-0.1cm}
% \caption{(a), (b)}
% %\vspace{-0.2cm}
% \label{fig:cor}
% \end{figure}

\begin{table}
\caption{Datasets and plan of experiments.}
%\vspace{-0.07cm}
\centering 

\begin{tabular}{
>{\centering\arraybackslash}m{0.2cm} 
>{\centering\arraybackslash}m{0.6cm} 
>{\centering\arraybackslash}m{3cm}
>{\centering\arraybackslash}m{3cm}} %>{\centering\arraybackslash}m{1.7cm}}
\hline
 &\textbf{Exp.} &\textbf{services} & \textbf{Applications} \\
\hline % inserts single horizontal line
%\multirow{1}{4em}
\rotatebox[origin=c]{90}{\textbf{Dataset I} } & \multirow{1}{4em}{\rotatebox[origin=c]{0}{exp. 1}}& VR Game, VR Video, VR chat/VoIP, AR, MR & Dirt Rally 2.0, Bigscreen, VR Chat, Solar System, Reality Mixer\\%\hline
\hline
\hline
{\rotatebox[origin=c]{90}{\textbf{Dataset II} }} & \multirow{1}{4em}{\rotatebox[origin=c]{0}{exp. 1}}& Slow VR Traffic, Fast VR Traffic & Steam VR Home, Beat Saber\\%\hline
\hline
\hline
\multirow{2}{4em}{\rotatebox[origin=c]{90}{\textbf{Dataset III}}} & \multirow{1}{4em}{\rotatebox[origin=c]{0}{exp. 1}}& Fast VR Traffic Game 1, Fast VR Traffic Game 2, Slow VR Traffic Game 1, Slow VR Traffic Game 2  & BeatSaber, Medal of Honor, Forklift Sim, Cooking Sim.\\ \cline{2-4}
%\hline
&  \multirow{1}{4em}{\rotatebox[origin=c]{0}{exp. 2}}& Slow VR Traffic, Fast VR Traffic  & SForklift Sim, Cooking Sim, BeatSaber, Medal of Honor\\%\hline

\hline
\end{tabular}
%\vspace{-0.4cm}
\label{table:exp}
\end{table}

\subsection{Performancce Metrics}
The performance of Metaverse network traffic prediction models can be evaluated using the following metrics: RMSE (Root Mean Squared Error), MAPE (Mean Absolute Percentage Error), and SMAPE (Symmetric Mean Absolute Percentage Error). These metrics quantify the differences between predicted and actual traffic values, aiding in assessing the model's accuracy. The RMSE measures the square root of the average squared differences between predicted and actual values, emphasizing more significant errors, and is given by:

$$
RMSE = \sqrt{\frac{1}{n} \sum_{i=1}^{n} \left( \hat{y}_i - y_i \right)^2, }
$$

where $\hat{y}_i$ represents the predicted value, $ y_i $ is the actual value, and $ n $ is the number of predictions. The MAPE is used to compute the average percentage error, offering an intuitive interpretation of prediction errors, and is formulated as:

$$
MAPE = \frac{100\%}{n} \sum_{i=1}^{n} \left| \frac{y_i - \hat{y}_i}{y_i} \right|.
$$

%This metric is useful when you want to understand the errors in percentage terms. 
Lastly, SMAPE provides a symmetric approach to percentage error by accounting for both over- and under-predictions. It is computed as:

$$
SMAPE = \frac{1}{n} \sum_{i=1}^{n} \frac{|\hat{y}_i - y_i|}{\left( |\hat{y}_i| + |y_i| \right)/2}.
$$

SMAPE balances the errors across different magnitudes of actual and predicted values, making it particularly suitable for dynamic environments like Metaverse traffic prediction. These metrics offer a comprehensive view of model performance, helping improve prediction accuracy for fluctuating network demands.

\subsection{Performance Evaluation}

The Tables \ref{tab:p1} \ref{tab:p2}, \ref{tab:p3}, \& \ref{tab:p4} present the performance evaluation of various models (Transformer, LSTM, GRU, and Stacked LSTM) across three datasets, with metrics such as RMSE, MAPE, and SMAPE. Table~\ref{tab:p1} shows the performance for frame size (Dataset I), Table~\ref{tab:p2} for frame count (Dataset II), Table~\ref{tab:p3} for frame IAT (Dataset III, exp 1), and Table \ref{tab:p4} for frame size (Dataset III, exp2). Each table compares the models' non-residual version (where residuals are not learned) with the residual learning approach, the ResLearn algorithm. In each case, SMAPE improvement is calculated, highlighting the percentage improvement in predictive accuracy when residual learning is applied.

The results indicate that the ResLearn algorithm significantly improves performance across all models and datasets, especially regarding SMAPE. For example, the transformer model in Table~\ref{tab:p1} achieves a SMAPE reduction from 0.78 to 0.24 (68.87\% improvement), and a similar trend is observed in Tables~\ref{tab:p2} and~\ref{tab:p3}, where the transformer and Stacked LSTM models show substantial SMAPE improvements of over 70\%. The observation is similar to Dataset III exp 2; however, GRU is better than the transformer. This demonstrates that residual learning can enhance the accuracy of time series models, particularly for the transformer architecture, making it the most effective among the evaluated models across all datasets.

\begin{table}
\centering
\caption{Performance for frame size from Dataset I}
\small % Reduce font size for table
\begin{adjustbox}{width=\columnwidth} % Fit within column width
\begin{tabular}{|l|c|c|c|c|}
\hline
\multirow{2}{*}{\textbf{Model}} & \multicolumn{3}{c|}{\textbf{Metrics}} &  \\
\cline{2-4}
 & \textbf{RMSE} & \textbf{MAPE} & \textbf{SMAPE} & \\
\hline
\multicolumn{5}{|c|}{\textbf{Non-Residual Algorithm}} \\
\hline
\textbf{Transformer} & \textbf{4872.49} & \textbf{0.0078} & \textbf{0.78} &  \multirow{4}{*}{\textbf{\% SMAPE Improvement}}\\
\textbf{LSTM}        & 4904.88 & 0.0079 & 0.79 &  \\
\textbf{GRU}         & 4915.61 & 0.0079 & 0.79 &  \\
\textbf{Stacked LSTM} & 4929.88 & 0.0079 & 0.79 & \\
\hline
\multicolumn{5}{|c|}{\textbf{ResLearn Solution}} \\
\hline
\textbf{Transformer} & \textbf{2164.95} & \textbf{0.0024} & \textbf{0.24} & \multirow{1}{*}{\textbf{68.87\%}} \\
\textbf{LSTM}        & 2725.03 & 0.0034 & 0.34 & \multirow{1}{*}{56.78\%} \\
\textbf{GRU}         & 2786.97 & 0.0031 & 0.31 & \multirow{1}{*}{61.04\%}\\
\textbf{Stacked LSTM} & 3666.30 & 0.0046 & 0.46 & \multirow{1}{*}{41.78\%} \\
\hline
\end{tabular}
\end{adjustbox}
\begin{tablenotes}
\item Note: %The metrics given for frame size prediction for Dataset I exp 1, refer to Table\ref{table:exp}. 
The top table provides results for the Non-residual version of the model in which residuals are not learned. The bottom table provides the result of the proposed solution. The \% SMAPE improvement is given in the fifth column.
\end{tablenotes}
\label{tab:p1}
%\vspace{-0.9cm}
\end{table}

\begin{table}
\centering
\caption{Performance for frame count from Dataset II}
\small % Reduce font size for table
\begin{adjustbox}{width=\columnwidth} % Fit within column width
\begin{tabular}{|l|c|c|c|c|}
\hline
\multirow{2}{*}{\textbf{Model}} & \multicolumn{3}{c|}{\textbf{Metrics}} &  \\
\cline{2-4}
 & \textbf{RMSE} & \textbf{MAPE} & \textbf{SMAPE} & \\
\hline
\multicolumn{5}{|c|}{\textbf{Non-Residual Algorithm}} \\
\hline
\textbf{Transformer} & \textbf{0.042} & \textbf{0.0072} & \textbf{0.72} &  \multirow{4}{*}{\textbf{\% SMAPE Improvement}}\\
\textbf{LSTM}        & 0.042 & 0.0073 & 0.73 &  \\
\textbf{GRU}         & 0.043 & 0.0076 & 0.76 &  \\
\textbf{Stacked LSTM} & 0.043 & 0.0076 & 0.77 & \\
\hline
\multicolumn{5}{|c|}{\textbf{ResLearn Solution}} \\
\hline
\textbf{Transformer} & \textbf{0.018} & \textbf{0.0020} & \textbf{0.20} & \multirow{1}{*}{\textbf{71.85\%}} \\
\textbf{LSTM}        & 0.017& 0.0023 & 0.23 & \multirow{1}{*}{68.67\%} \\
\textbf{GRU}         & 0.018 & 0.0025 & 0.25 & \multirow{1}{*}{67.09\%}\\
\textbf{Stacked LSTM} & 0.019 & 0.0022 & 0.22 & \multirow{1}{*}{71.72\%} \\
\hline
\end{tabular}
\end{adjustbox}
% \begin{tablenotes}
% \item Note: The metrics given for frame size prediction for Dataset I exp 1, refer to Table\ref{table:exp}. The top table provides results for the Non-residual version of the model in which residuals are not learned. The bottom table provides the result of the proposed solution. The \% SMAPE improvement is given in the fifth column.
% \end{tablenotes}
\label{tab:p2}
%\vspace{-0.9cm}
\end{table}

\begin{table}
\centering
\caption{Performance for frame IAT from Dataset III exp. 1}
\small % Reduce font size for table
\begin{adjustbox}{width=\columnwidth} % Fit within column width
\begin{tabular}{|l|c|c|c|c|}
\hline
\multirow{2}{*}{\textbf{Model}} & \multicolumn{3}{c|}{\textbf{Metrics}} &  \\
\cline{2-4}
 & \textbf{RMSE} & \textbf{MAPE} & \textbf{SMAPE} & \\
\hline
\multicolumn{5}{|c|}{\textbf{Non-Residual Algorithm}} \\
\hline
\textbf{Transformer} & \textbf{0.037} & \textbf{0.0032} & \textbf{0.32} &  \multirow{4}{*}{\textbf{\% SMAPE Improvement}}\\
\textbf{LSTM}        & 0.034 & 0.0029 & 0.29 &  \\
\textbf{GRU}         & 0.034 & 0.0030 & 0.30 &  \\
\textbf{Stacked LSTM} &0.033 & 0.0027 & 0.29 & \\
\hline
\multicolumn{5}{|c|}{\textbf{ResLearn Solution}} \\
\hline
\textbf{Transformer} & \textbf{0.029} & \textbf{0.0020} & \textbf{0.20} & \multirow{1}{*}{\textbf{38.01\%}} \\
\textbf{LSTM}        & 0.030& 0.0021 & 0.21 & \multirow{1}{*}{25.11\%} \\
\textbf{GRU}         & 0.035 & 0.0025 & 0.25 & \multirow{1}{*}{13.74\%}\\
\textbf{Stacked LSTM} & 0.028 & 0.0019 & 0.19 & \multirow{1}{*}{36.04\%} \\
\hline
\end{tabular}
\end{adjustbox}
% \begin{tablenotes}
% \item Note: The metrics given for frame size prediction for Dataset I exp 1, refer to Table\ref{table:exp}. The top table provides results for the Non-residual version of the model in which residuals are not learned. The bottom table provides the result of the proposed solution. The \% SMAPE improvement is given in the fifth column.
% \end{tablenotes}
\label{tab:p3}
%\vspace{-0.9cm}
\end{table}

\begin{table}
\centering
\caption{Performance for frame size from Dataset III exp. 2}
\small % Reduce font size for table
\begin{adjustbox}{width=\columnwidth} % Fit within column width
\begin{tabular}{|l|c|c|c|c|}
\hline
\multirow{2}{*}{\textbf{Model}} & \multicolumn{3}{c|}{\textbf{Metrics}} &  \\
\cline{2-4}
 & \textbf{RMSE} & \textbf{MAPE} & \textbf{SMAPE} & \\
\hline
\multicolumn{5}{|c|}{\textbf{Non-Residual Algorithm}} \\
\hline
\textbf{Transformer} & {78.409} & {0.0103} & {1.04} &  \multirow{4}{*}{{\% SMAPE Improvement}}\\
\textbf{LSTM}        & 74.140 & 0.009 & 0.98 &  \\
\textbf{GRU}         & \textbf{78.014} & \textbf{0.0103} & \textbf{1.03} &  \\
\textbf{Stacked LSTM} & 78.677 & 0.0104 & 1.04 & \\
\hline
\multicolumn{5}{|c|}{\textbf{ResLearn Solution}} \\
\hline
\textbf{Transformer} & {42.950} & {0.0041} & {0.41} & \multirow{1}{*}{\textbf{60.69\%}} \\
\textbf{LSTM}        & 42.860& 0.0041 & 0.41 & \multirow{1}{*}{58.08\%} \\
\textbf{GRU}         & \textbf{40.724} & \textbf{0.004} & \textbf{0.40} & \multirow{1}{*}{61.39\%}\\
\textbf{Stacked LSTM} & 48.184 & 0.0042 & 0.42 & \multirow{1}{*}{60.1\%} \\
\hline
\end{tabular}
\end{adjustbox}
% \begin{tablenotes}
% \item Note: The metrics given for frame size prediction for Dataset I exp 1, refer to Table\ref{table:exp}. The top table provides results for the Non-residual version of the model in which residuals are not learned. The bottom table provides the result of the proposed solution. The \% SMAPE improvement is given in the fifth column.
% \end{tablenotes}
\label{tab:p4}
%\vspace{-0.1cm}
\end{table}

\subsection{Performance Comparision and Discussion}

\begin{figure}
        \centering
        \begin{subfigure}{0.24\textwidth}
            \centering
            \includegraphics[width=\textwidth]{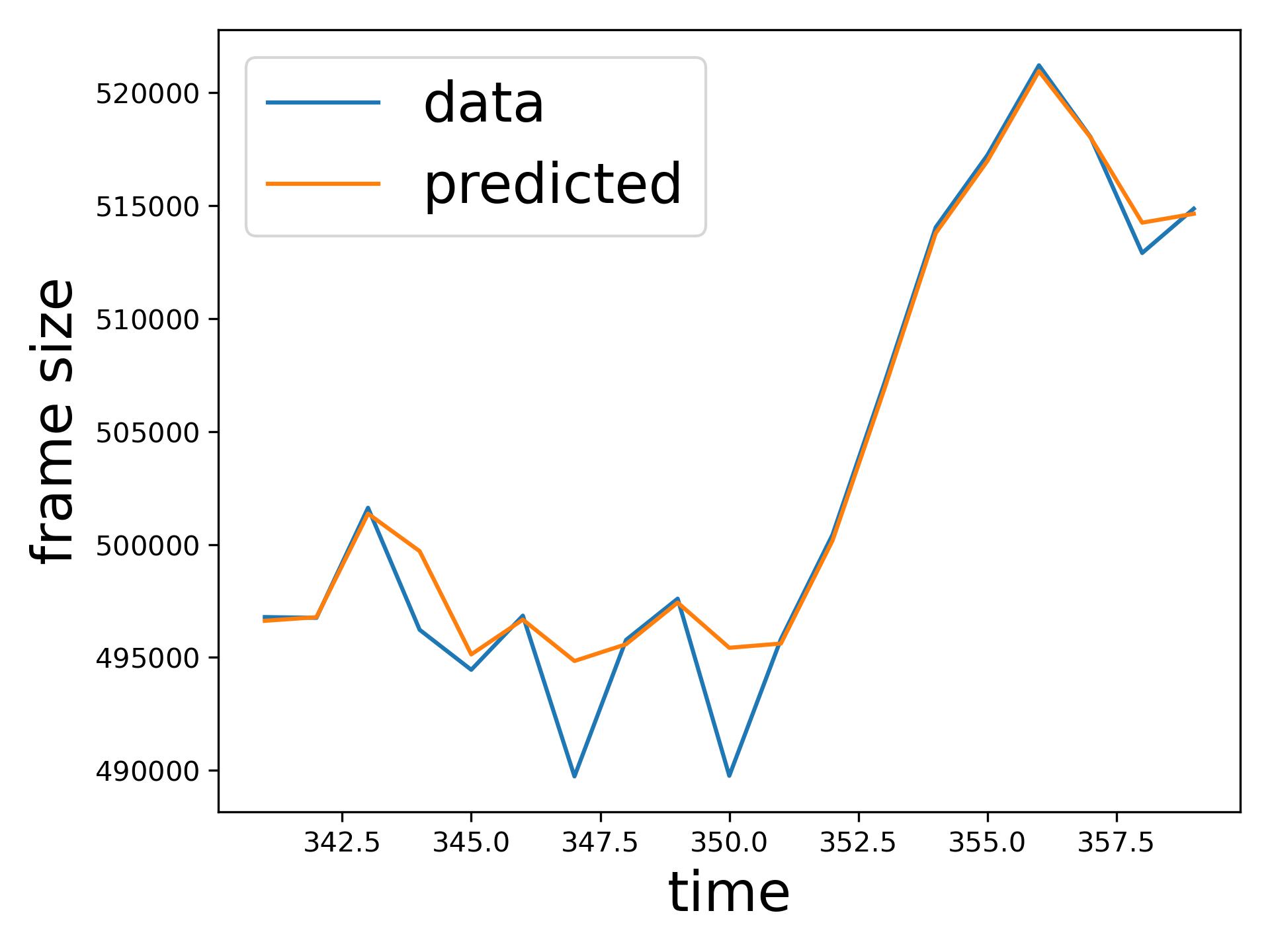}
            \caption{}
            %\label{fig:mean and std of net14}
        \end{subfigure}
        %\hfill
        \begin{subfigure}[b]{0.24\textwidth}  
            \centering 
            \includegraphics[width=\textwidth]{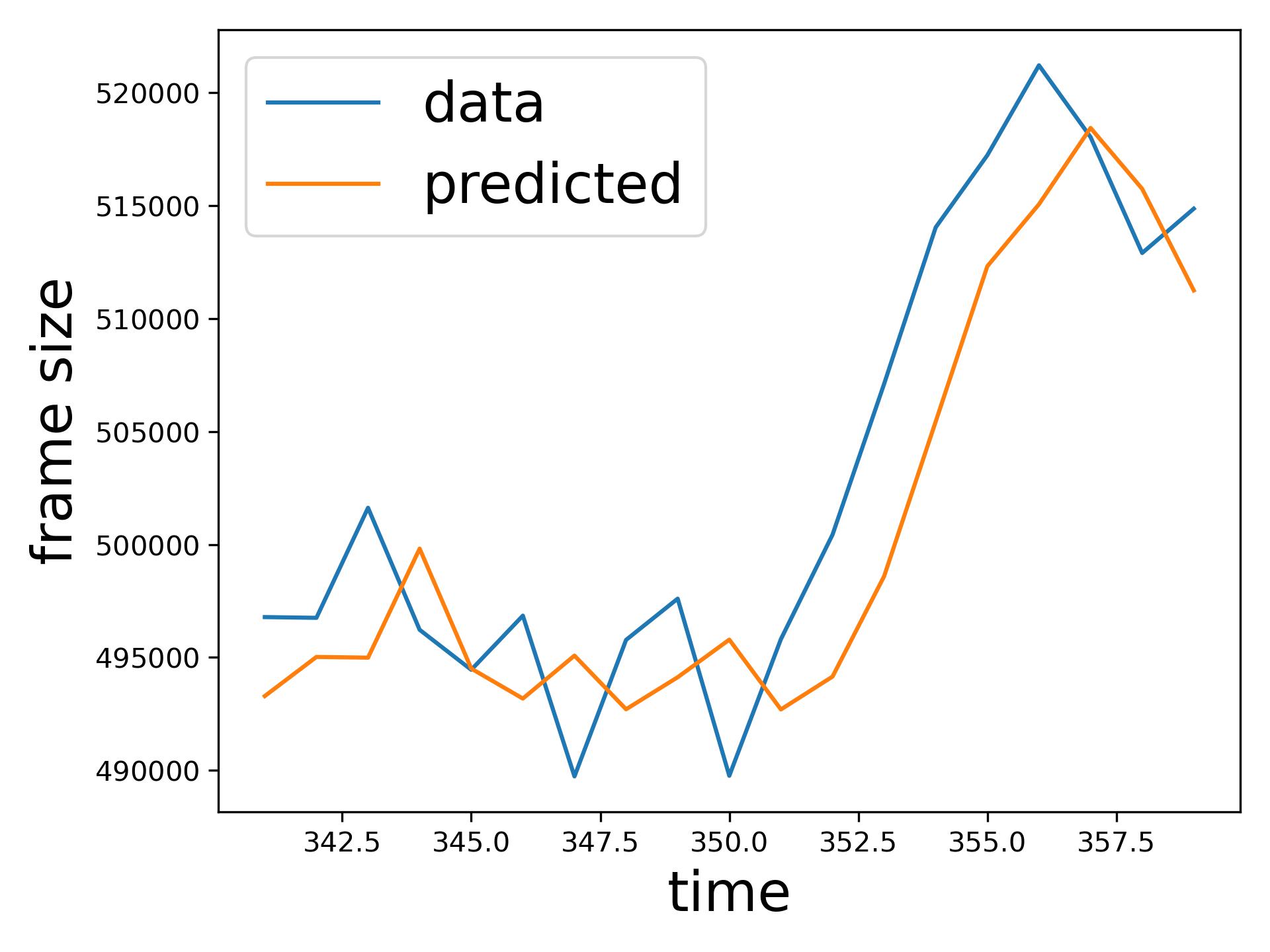}
            \caption{} 
            \label{fig:mean and std of net24}
        \end{subfigure}
        \vskip\baselineskip
        \begin{subfigure}[b]{0.24\textwidth}   
            \centering 
            \includegraphics[width=\textwidth]{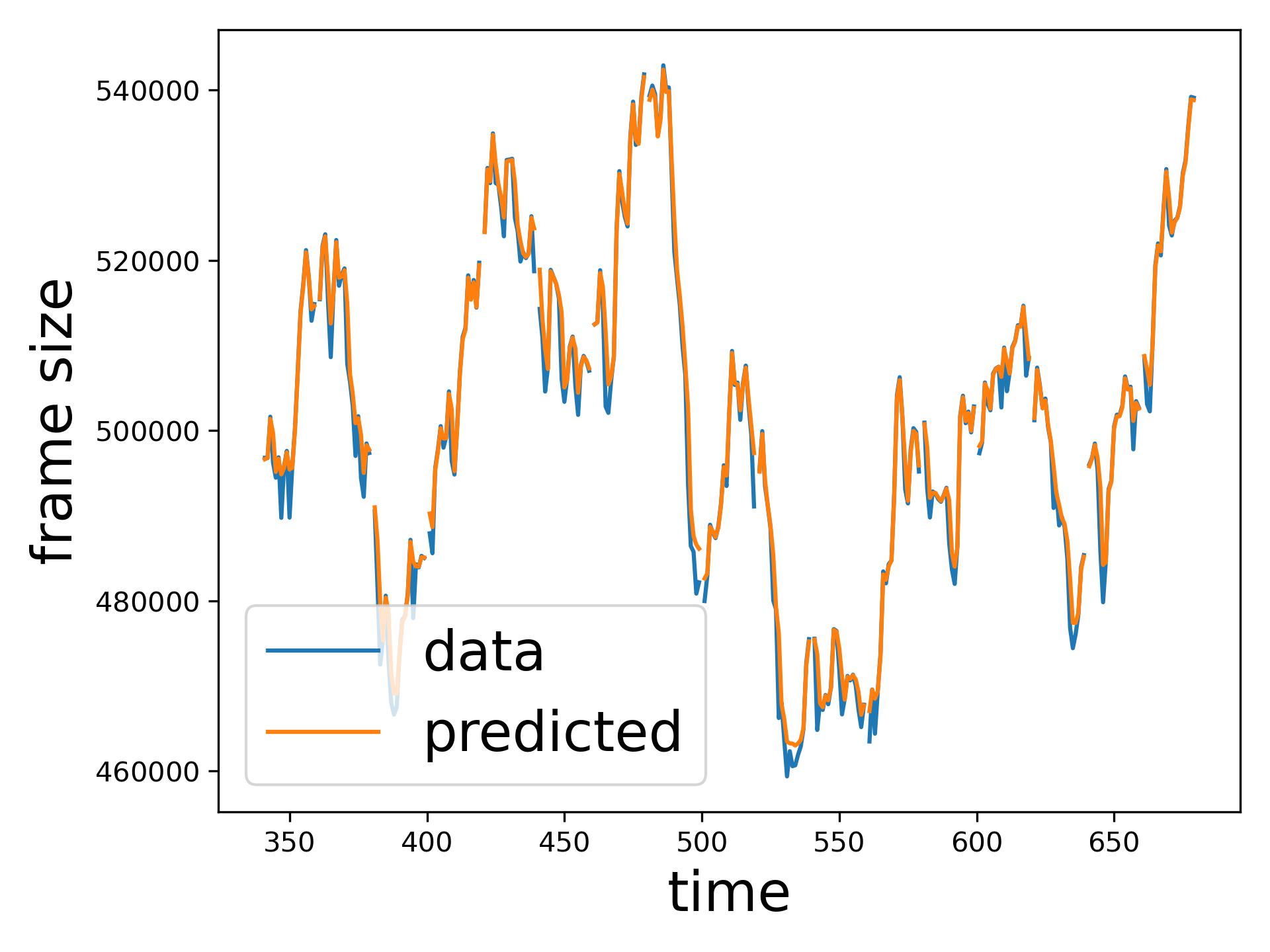}
            \caption[]%
            {{\small }}    
            \label{fig:mean and std of net34}
        \end{subfigure}
        \hfill
        \begin{subfigure}[b]{0.24\textwidth}   
            \centering 
            \includegraphics[width=\textwidth]{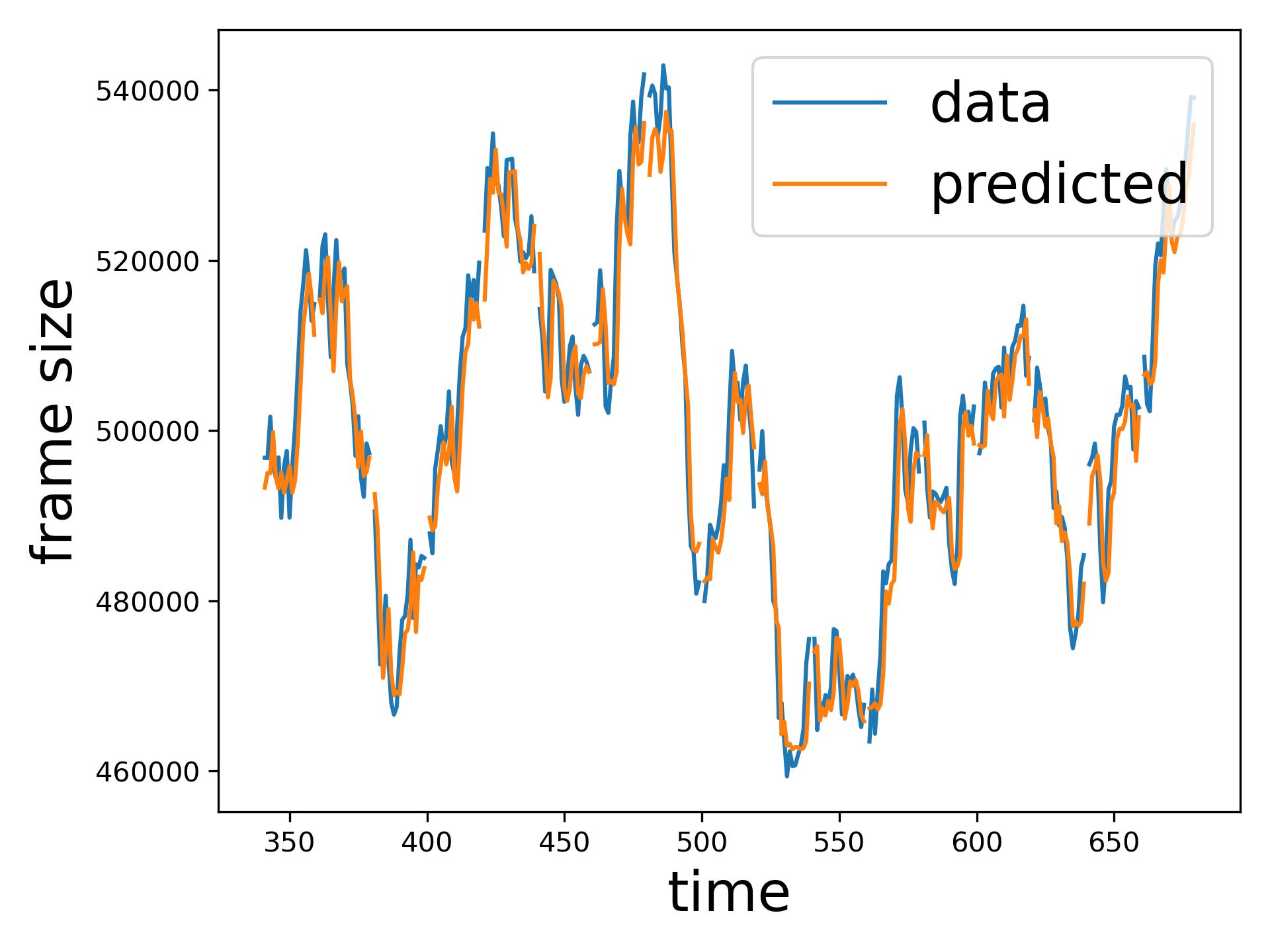}
            \caption[]%
            {{\small }}    
            \label{fig:mean and std of net44}
        \end{subfigure}
        \caption{(a) ResLearn prediction of 18th segment for Dataset I, (b) 18th segment prediction from non-ResLearn solution, (c) ResLearn's Overall prediction of Dataset I, (d) non-ResLearn's overall prediction of Dataset I. All four figures are predicted for frame size. ResLearn is significantly better at predicting the peak necessary for network management for higher resource consumption.} 
        %\vspace{-0.7cm}
        \label{fig:resc}
    \end{figure}

The comparison of SMAPE between the SoA transfer learning model \cite{10437897} and the proposed ResLearn solution (Table~\ref{table:pc}) demonstrates a significant performance improvement in favour of ResLearn. In various traffic conditions, as considered in \cite{10437897}, such as BeatSaber and Steam VR house at different Mbps rates, ResLearn consistently achieves a near-perfect reduction in SMAPE, reaching over 99\% improvement across all scenarios. %For example, in the BeatSaber 40 Mbps case, the SMAPE dropped from 404.05 in the SoA model to just 0.36 in ResLearn, a 99.91\% improvement. 
Similar reductions are observed in other settings, like the Steam VR house 40 Mbps, where SMAPE decreased from 562.87 to 0.45, showcasing ResLearn's superior accuracy. Figure \ref{fig:resc} compares time series prediction of ResLearn and non-ResLearn solutions. Overall, the ResLearn solution is superior for network management because it accurately predicts the peaks at which maximum resource is required.

%Overall, ResLearn outperforms the transfer learning SoA in every scenario tested, achieving nearly flawless prediction accuracy with SMAPE improvements of at least 99.64\% across the board. This remarkable difference highlights the effectiveness of the residual learning approach employed by ResLearn, allowing it to handle a wide range of traffic conditions far more efficiently than the state-of-the-art transfer learning method. The drastic reduction in error rates underscores ResLearn's potential for use in complex traffic prediction tasks where high accuracy is crucial.

\begin{table}
\centering
%\scriptsize
\caption{SMAPE comparison between Transfer Learning SoA and ResLearn Solution}
\begin{tabular}{
>{\centering\arraybackslash}m{2.5cm} 
>{\centering\arraybackslash}m{2cm} 
>{\centering\arraybackslash}m{1cm}
>{\centering\arraybackslash}m{1.5cm}}
\hline
\textbf{Traffic} & \textbf{Transfer Learning SoA \cite{10437897}} & \textbf{ResLearn}  & \textbf{\% SMAPE Improvement}\\ \hline
\multirow{1}{*}{\rotatebox[origin=c]{0}{BeatSaber 40 Mbps}} & 404.05 & 0.36 & \textbf{99.91\%}\\ \hline
\multirow{1}{*}{\rotatebox[origin=c]{0}{BeatSaber 54 Mbps}} & 285.29 & 1.01 & \textbf{99.64\%}\\ \hline
\multirow{1}{*}{\rotatebox[origin=c]{0}{BeatSaber 120 Mbps}} & 371.82 & 0.15 & \textbf{99.95\%}\\ \hline
\multirow{1}{*}{\rotatebox[origin=c]{0}{Steam VR house 40 Mbps}} & 562.87 & 0.45 & \textbf{99.92\%}\\ \hline
\multirow{1}{*}{\rotatebox[origin=c]{0}{Steam VR house 54 Mbps}} & 404.41 & 0.69 & \textbf{99.82\%}\\ \hline
\end{tabular}
\label{table:pc}
\end{table}

%\subsection{Discussions}

% \begin{figure}
% %\vspace{-0.4cm}
% \title{}
% \centering
% \subfloat[]{%
%   \includegraphics[width=0.8\linewidth]{rect1029.png}%
%   \label{fig:sm_1}
% }
% \vspace{-0.01cm}
% \subfloat[]{
% \includegraphics[trim=0 0 0.3cm 0 ,clip, width=0.8\linewidth]{rect1120.png}%
% \label{fig:sm_2}
% }
% %\vspace{0.1cm}
% \vspace{-0.1cm}
% \caption{(a) System model of the proposed solution, and (b) Residual Learning (ResLearn) algorithm's training process.}
% %\vspace{-1cm}
% \label{fig:prd}
% \end{figure}

\section{Conclusion and Future Work}

Our work significantly advances Metaverse network traffic prediction by introducing a comprehensive, real-world dataset and developing novel algorithms including the view-frame and ResLearn algorithms. These algorithms enable ISPs to manage network resources in an effective manner, satisfying the QoS and enhancing the user experience for Metaverse applications. Given that our solution substantially reduces prediction errors about 99\% than the SoA \cite{10437897}, future work can focus on expanding the dataset to cover a broader range of Metaverse applications and environments, integrating advanced AI techniques to improve prediction accuracy further, and exploring real-time deployment in diverse network architectures. Additionally, adaptive algorithms for dynamic resource allocation in response to traffic fluctuations will be investigated for enhancing the robustness in provisioning Metaverse ecosystem.

\bibliographystyle{IEEEtran}
\bibliography{refs}

\end{document}